# Indian Commercial Truck License Plate Detection and Recognition for Weighbridge Automation

Siddharth Agrawal[1] and Keyur D. Joshi[2]

*Abstract*— Detection and recognition of a licence plate is important when automating weighbridge services. While many large databases are available for Latin and Chinese alphanumeric license plates, data for Indian License Plates is inadequate. In particular, databases of Indian commercial truck license plates are inadequate, despite the fact that commercial vehicle license plate recognition plays a profound role in terms of logistics management and weighbridge automation. Moreover, models to recognise license plates are not effectively able to generalise to such data due to its challenging nature, and due to the abundant frequency of handwritten license plates, leading to usage of diverse font styles. Thus, a database and effective models to recognise and detect such license plates are crucial. This paper provides a database on commercial truck license plates, and using state-of-the-art models in real-time Object Detection: You Only Look Once Version 7, and Scene Text Recognition: Permuted Autoregressive Sequence Models, our method outperforms the other cited references where the maximum accuracy obtained was less than 90%, while we have achieved 95.82% accuracy in our algorithm implementation on the presented challenging license plate dataset.

## I. INTRODUCTION

### A. Background Literature

In License Plate Recognition (LPR), [1] used a Convolutional Neural Network + Bi-LSTM model to extract the bidirectional 1D spatial attention maps to attend to each character region at a time. SVM + ANN solutions have also been shown to have good accuracy for classification [2]. The current state-of-the-art for LPR, [3] used a BiseNet[4] to compute the position-wise and character recognition segmentation maps of characters in parallel using semantic segmentation, and then combined these segmentation maps using a shared classifier.

The recent developments in Scene Text Recognition (STR) have been towards combining visual and language modelling to achieve better accuracy [5], [6], [7], [8]. [9] developed ABINet to increase the accuracy of visual recognition by incorporating a language model. They first compute the character probability predictions using a vision recognition model and then feed these into a language model. The cloze masking strategy was developed in order to obtain bidirectional feature representations. The cloze mask strategy proposed in their Bidirectional Cloze Network (BCN) worked by employing attention masking along the diagonal and using a multi-headed cross attention transformer to avoid leaking information across time steps and transformer layers. This masks a single position's predictions at every timestep during training, and the model must predict this position's character probability labels in order to improve its loss. Thus, leading to more effective, bidirectional language comprehension. ABINet then uses the gated mechanism in order to generate predictions. These predictions are then iteratively corrected by feeding the outputs into the Language Model repeatedly, thus, improving prediction accuracy by coping with noisy contexts using language cues, and also alleviating the misalignments in sequence length, which is often an issue in parallel decoding models, by fusing visual and linguistic features multiple times. While ABINet used language modelling as a form of "spell check", such modelling would sometimes result in incorrect corrections made to otherwise correct predictions made by the VRM, resulting in lower accuracy. Recent progress [5], [6] is aimed towards combining an ILM and a VRM in order to maximize the joint probability of accurate predictions, instead of independently modelling both. While ABINet modelled language and visual representations independently, [5] introduced a model to optimise the joint probability of correct labels given an image. They extracted character recognition labels using a segmentation-based model, and then by using Graph Textual Reasoning (GTR) model and language modelling to produce output labels in parallel, with a consistency loss between the GTR and language modelling based predictions, they optimised the joint probability of correct labels given an image. ViTSTR [10] were the first ones to utilise a Vision Transformer for STR.

A vision transformer [11] is an effective backbone for LPR as well, as it has a fast inference speed (9.8 milliseconds) as it has a far lower depth as compared to Convolutional Neural Networks, and thus more operations can be done in parallel on a GPU. Furthermore, this also allows it to decode characters parallelly, as opposed to autoregressively. Autoregressive decoding can reduce inference speed considerably as the characters are decoded one by one and not all at once. [12] further showed that, in neural machine translation, such a parallel decoding scheme can maintain competitive accuracy with autoregressive decoding by implementing certain optimisations. Later, [13] used a Swin Transformer [14] backbone to capture hierarchical transformer feature representations using shifted windows, improving accuracy.

*This work was supported by Ahmedabad University
[1]Siddharth Agrawal is an Undergraduate Student, School of Arts and Sciences, Ahmadabad University, Ahmedabad, India siddharth.a@ahduni.edu.in
[2]Keyur D. Joshi is an Assistant Professor, School of Engineer- ing and Applied Sciences, Ahmadabad University, Ahmedabad, India keyur.joshi@ahduni.edu.in

## B. Information about PARSeq

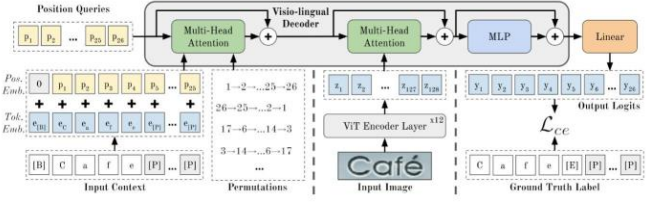

Fig. 1. Architecture of PARSeq

Parseq combines insights from ABINet, VITStr and XL-Net [15] to come up with the current state-of-the-art in STR. Furthermore, with recent developments in hardware and GPUs, it enables shallower networks with high parallelism such as transformers to produce faster and more accurate results, leading to higher accuracy and faster inference times, which makes it ideal for a task such as LPR, where inference speed has to be maximised. PARSeq iteratively refines the predictions using an Ensemble of Language Modelling models, wherein, it trains using an ensemble of language modelling methods with shared weights, and in inference, can either decode parallelly or autoregressively. The predictions can be further refined using iterative refinement. ParSeq not only has the highest accuracy in the STR task, being the current state-of-the-art, but also has some of the lowest inference times on capable GPUs: 12 milliseconds for non-autoregressive decoding, and 15 milliseconds for autoregressive decoding on an NVIDIA Tesla A100 GPU.

While ViTSTR had no language modelling component, PARSEQ [6] attained state-of-the-art in STR by combining the features from a Permutated Language Modelling (PLM) multi-head attention model, with the encoded features from a Visual Transformer (ViT) backbone. Not only is the ViT state-of-the-art for many vision tasks, but it also has a relatively fast inference speed, especially on small images. With the smaller patch embeddings required for smaller images such as license plates, it is also memory efficient.

Fig. 2. PARSeq as an ensemble of various language decoding techniques

While a ViT without any language modelling or iterative refinement may suffice for simpler license plates with smaller text, such as ones found in CCPD [16] or AOLP [17] databases, it is not sufficient for longer license plates with restricted formats such as the Indian license plates. While License Plates do not have a language similar to natural languages such as English, language modelling can greatly benefit in implicitly enforcing format, even in cases where postprocessing the labels with explicit format restrictions may fail, and enhances positional and visual features in our testing, leading to the model being less prone to visual misalignments.

## C. Existing Databases

Existing databases such as CCPD provide a large number of images. However, simply training on the CCPD database alone proved insufficient to achieve satisfactory accuracy on Indian license plate databases in testing. These databases have a smaller license plate text sequence length, and do not have any two-line examples, it is insufficient to generalise models trained on these databases to the Indian context and License Plates. While other databases exist, they fail to generalise to the long sequence length and double or triple-line formats of the Indian license plates.

Many papers such as [18], [19], and [6], outline the increased efficacy of real-world STR databases as compared to synthetic database. Thus, we make a similar claim that synthetic databases of Indian license plates alone would not generalise well to real-world license plates. Thus, arising the need for such a database with samples from the real world.

While some Indian license plate databases exist such as [20], [21], [22], [23], [24], they largely focus on samples that are clearer and free of distortions. These do not have a diversity of fonts, are either small-scale or not publicly available, and are not focused on challenging examples on commercial license plates such as the prepared database. Furthermore, the proposed method utilises more recent state-of-the-art techniques in Computer Vision such as Transformers and YOLOv7 to achieve high inference speed and accuracy on challenging samples, even in multi-line text samples. While some synthetic text databases and small-scale databases are publically available, they are unable to achieve high accuracy by themselves, due to their inability to generalise to real-world testing samples in our experiments. Thus, a database on real Indian license plates was required.

## D. Challenges Related to Recognition of Indian License Plates

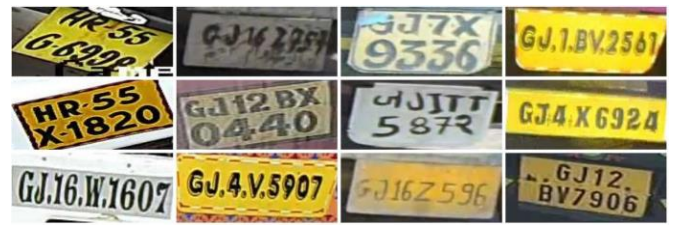

Fig. 3. Diversity of fonts in the database

License plate detection and its recognition play an important role in several real-world tasks such as traffic law enforcement, surveillance systems, toll booth systems, logistics, car parking registration management, etc. The prepared database as shown in the next subsection is primarily aimed at Indian license plates, which propose a difficult scenario for even the state-of-the-art in LPR due to several reasons. Rotation, Illumination and shear-based distortions, are common and well-studied, however, Indian license plates add further challenges such as 1) handwritten text or text of different fonts and formats, 2) dirt and damage to the number plate itself that obscures the text, 3) partially visible license plates due to the overhanging netting or grills at the back of the vehicle, and 4) longer and more complex sequence length which can cause lower accuracy even if the character level accuracy or Normalised Edit Distance (NED) is low. Based on the capturing device, the inbuilt JPEG compression factor also reduces the quality of the images, resulting in an average JPEG quality of 73% in this database. Another significant hurdle of the database is the fact that many license plates in the presented database are more than one line. Many samples are either two-line or three-line license plates. This can cause major visual misalignment issues when using positional embedding or transformer-based architectures, giving erroneous repetitions or deletions within the text sequence.

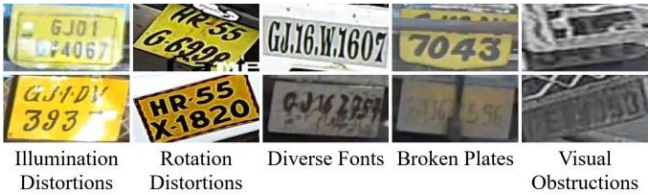

Fig. 4. Diversity of image conditions, challenging scenarios, and distortions

*E. Background on Database Collection*

Weighbridges or Weigh Stations are common in India for weighing Trucks containing cargo such as crops, water, raw materials, etc. It serves as a quick and easy way to automate the process of weighing large amounts of cargo so that it can be sold for the appropriate price. A few images of the license plate were acquired by such weighbridges, to keep accounts of the amount of goods sold by an owner. The license plate is recorded along with the weight and type of goods sold. This is how the database has been sourced. The weighbridges provided images from three states of India: 1) Gujarat, 2) Haryana and 3) Maharashtra. Figures X, Y, Z, A, B, C, and D provide a few samples of the various image categories.

On Indian trucks, license plate information can be usually found in three places. The front license plate, the rear license plate, and a license plate were painted on the back of the truck's tanker. All of these positions were annotated and labelled for added redundancy. If the LPR system using a single camera cannot detect or recognise a license plate due to some issue, then a system with two cameras, where one is at the back of the vehicle, and the other is at the front, may still be able to recognise the license plates.

## II. METHODS

*A. Database Annotation*

Each Image within this database is labelled using a multi-point polygon around the license plate boundary and an associated bounding box. A wide range of annotation categories are used to effectively differentiate the diverse database:
- *partial_text*: The text on the license plate is only partially visible (fig. 5 orange-coloured boxes). The reasons

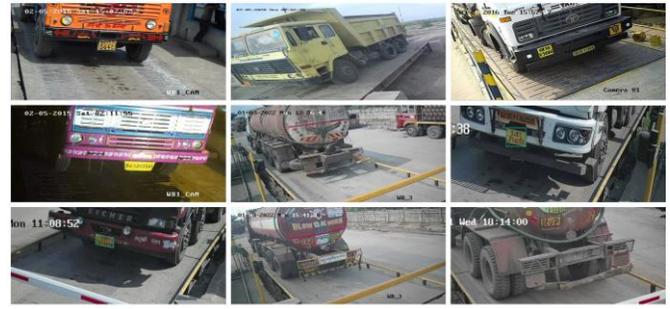

Fig. 5. Samples to visualise annotations

  may include dirt, illumination distortions, reflections, broken license plate or visual obstructions, etc.
- *obscured*: The text on the license plate is only partially visible due to some form of visual obstruction (fig. 5 orange-coloured boxes).
- *unreadable*: The text on the license plate is unreadable (fig. 5 red coloured boxes). The reasons may include dirt, low resolution, etc.
- *barely_readable*: Text that is barely readable even to a human. Extremely challenging and unclear. Therefore, such samples serve as a good and diverse database for training, but are difficult to recognise even by humans, and therefore, were removed from the validation split.
- *double_plate*: In very limited scenarios, a bounding box may contain two different license plates on top of each other. This category is used to denote such scenarios. Such an annotation likely contains two more bounding box annotations of the individual license plates within its bounding box.
- *license_plate*: A license plate with none of the aforementioned defects or eccentricities.

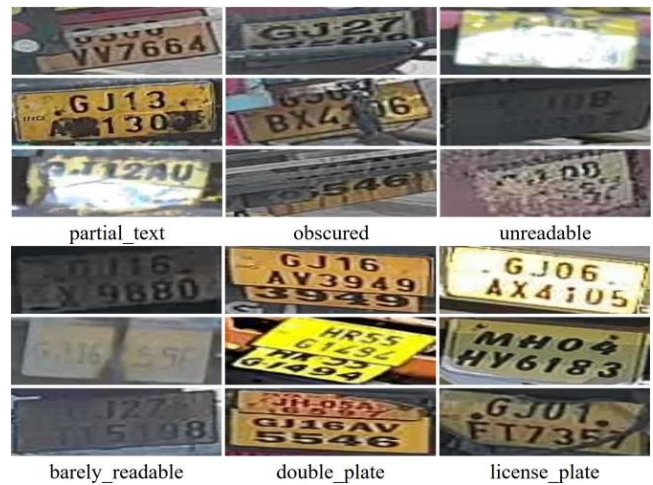

Fig. 6. Illustration of license plate annotation categories

License plate annotation categories are not mutually exclusive, i.e., a single license plate annotation can belong to multiple of the aforementioned categories. This is especially the case with categories *partial_text* and *obscured*, as incomplete text is often the result of some form of visual obstruction. However, this is not always the case, as faded-out text, dirt, etc., can also cause certain characters to be illegible (as seen in fig. 6), thus placing such plates into the *partial_text* category. *Obscured* license plates usually result from some grill, net, mud flaps, or other obstruction, that extends out of the rear part of the vehicle's trailer. All of these bounding boxes were used to train the object detection model.

All categories of license plates except *unreadable* were considered as a single class while training for license plate detection and recognition as it was hypothesized to provide improved generalizability, based on preliminary testing. For the validation split of LPR, only the *license_plate* category was used, i.e., license plates with none of the aforementioned defects or eccentricities.

Many of the license plates may be repeated in a database due to two reasons. First, the cameras were placed in such a fashion, that multiple points of recognition can be used. Both the front and the back of the vehicle can be scanned for redundancy and more effective LPR. Secondly, many vehicles pass through the weighbridges or cameras multiple times throughout the year, and therefore, their license plates are captured multiple times as well.

While many license plate examples follow a standardised font specified by the Regional Transport Office(s) (RTO) of India, many commercial vehicles in India are not mandated to follow this standardised font as seen in Fig. 4. Thus, commercial Indian license plates are found in a plethora of fonts of characters. This also makes it more challenging to train a model that results in effective accuracy.

### B. Synthetic Database

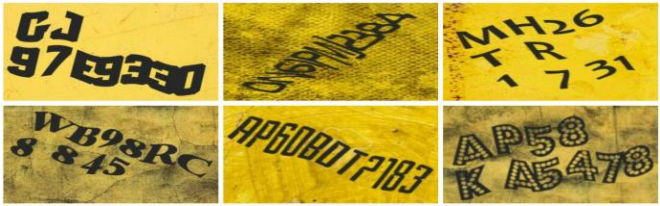

Fig. 7. Samples from the generated synthetic database

Synthetic license plates for various states and districts were generated using an image of a yellow background, over which text from one of 8445 fonts was used. Newline characters were also introduced selectively after either the state code, RTO series code, or the district code, with a probability of 50%, and an added probability of a new-line character present at any of these three places with a probability of 5%, to mimic multi-line text license plates. Examples of the generated synthetic license plates are shown in fig. 7. These synthetic databases also help to enforce language modelling of license plate state and district codes that were either minority classes, or missing, in the presented real license plates database. In the absence of these, the model is more prone to recognising erroneous state or district codes, and is biased towards the majority class, due to an imbalance in the database's regional codes. This allowed for more generalisability and diversity to various fonts, and state and district codes. TextRecognitionDataGenerator (TRDG) [25], with some modifications was used to generate the synthetic samples, and several augmentations were added to the synthetic database:

- Random blur of radius 0 to 4.
- Random rotation with an angle of -90° to +90°.
- Either none, sine, or cosine text warping, all with equal probability.

Further augmentations to both the real and synthetic databases were applied during training, which are mentioned in the next section.

### C. Augmentations

Typical STR-based strategies would approach the problem by decomposing this problem into two subtasks: text detection, and text recognition. First, in text detection, individual lines of text are detected and cropped. Either segmentation polygons or bounding boxes are returned. Then, the recognition model runs inference for each line of text. Finally, the labels are concatenated for the final output. As compared to the car license plates, commercial Indian licence plates pose a significant hurdle to recognition, and yet, are one of the most important classes of vehicles in need of an effective LPR model due to their numerous applications in logistics management and automation. Extreme care was taken to ensure that all repeated license plates were kept in the same split of the database, to avoid leakage of samples and information from the training split to the validation split. However, such a method significantly increases inference times as two separate models need to be run sequentially. While the recognition of multiple lines can be processed parallelly in batches, which helps alleviate the issue, such a method is still slower than inferencing a single recognition model.

## III. RESULTS AND DISCUSSION

As such a two-stage recognition method would considerably reduce inference speed, we chose to instead train the recognition model on augmented data that incorporates multi-line text. As the transformer self-attention mechanism is implemented on 2-dimensional patch embeddings, and spans both the width and the height of the image, it should be able to incorporate both vertical and horizontal feature alignments, and thus, effectively recognize complex layouts in 2-dimensional space, i.e., can recognise multi-line text samples effectively while maintaining accuracy on single-line text samples. The data augmentations are described below:

- Shear Distortions (percentage): ±0.9 in the horizontal direction and ±0.2 in the vertical direction
- Rotation Distortions: ±30°
- Translation (percentage): ±0.1 in the horizontal direction and ±0.3 in the vertical direction
- Additive Poisson Noise: lambda = Uniform distribution from 0 to 40 in discrete steps of 5
- Gaussian Blur: radius = Uniform distribution from 0 to 4 in discrete steps of 1

TABLE I
YOLOv7 Accuracy on the Presented Dataset

| Database | Resolution | Precision | Recall | mAP at 0 to 0.5 | mAP at 0.5 to 0.95 | maximum F1-Score | FPS |
|---|---|---|---|---|---|---|---|
| All included | 640x640 | 0.938 | 0.958 | 0.987 | 0.667 | 0.95 | 121 |
| Only fully readable | 640x640 | 0.991 | 0.995 | 0.997 | 0.702 | 0.99 | 121 |

Note: The maximum F1-score is the maximum recorded F1-score across several possible values of confidence thresholds for the final model, whereas the other metrics were recorded from the epoch with the highest validation mAP while training.

- Concatenating two images vertically: With a 50% probability, two images chosen at random are concatenated on top of each other (vertically) and their text labels are also concatenated. This helps increase accuracy on two-line LPR considerably, as seen in Table III. This strategy was only used for pretraining, and was only employed on non-license plate samples.
- Other augmentations as per [26]'s RandAugment strategy.

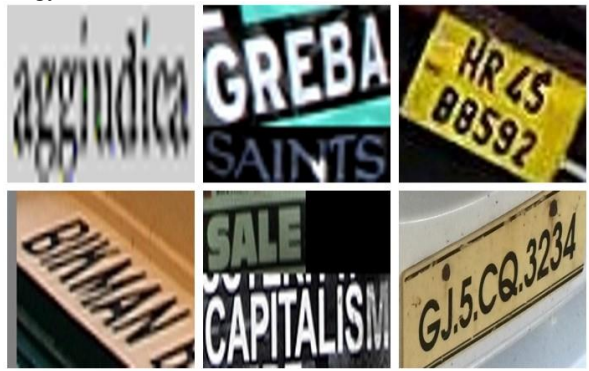

Fig. 8. Training Samples after Augmentations.

As [27] previously demonstrated in their paper, image concatenation is a valuable augmentation that may increase the out-of-vocabulary accuracy of STR models in unseen and unfamiliar contexts. The method allows for effective double-line LPR, even though STR models are primarily made and pretrained for single-line text recognition. This helps us reduce the inference time considerably as a text detector is not required to be incorporated into the pipeline. It also reduces errors caused due to erroneous text detections, which can propagate to the recognition step since there is no backpropagation of errors from the recognition to the text detection. Furthermore, due to the database being relatively small, synthetic databases and other large STR databases were incorporated into the pretraining.

First, we discuss the implementation details and results of the detection model, followed by the recognition model.

### A. License Plate Detection

License Plate Detection was trained on an NVIDIA Quadro P4000 GPU using a pretrained YOLOv7 [28] model as it is the current state-of-the-art in real-time object detection. The YOLOv7 base model was trained for 100 epochs with the adam optimizer with a batch size of 92, with the One Cycle Learning Rate Policy [29] with an initial learning rate of 0.01 and final learning rate of 0.2, using 2237 training images from the presented database, and was validated on 239 images from the presented database.

### B. License Plate Recognition

A slight modification to PARSeq was made. The patch embedding size of the conventional PARSeq was of size 4x8 with an image resolution of 32x128. While this works well for most STR and LPRtasks, it performs poorly on our database due to the limited vertical resolution, which impairs its performance in two-line and three-line LPR, being prone to misalignments, recognition errors, and erroneous repetition of characters. To effectively recognise the license plates of multiple lines, the model was modified to have an increased image resolution of 224x224 at the cost of a larger patch embedding size of 16x16, resulting in the same number of total parameters as the original model (24M).

TABLE II
Results of Changing Patch Size, Resolution, and Aspect Ratio

| Resolution | Patch size for embedding | Validation accuracy % | Validation NED | Params (M) |
|---|---|---|---|---|
| 38x128 | 4x8 | 58.96 | 89.72 | 24.4 |
| 224x224 | 16x16 | 73.13 | 95.70 | 24.4 |

The training set of 2237 images did not suffice in getting adequate results due to the diversity of character fonts and placements. The model was prone to visual misalignment and made errors of repeating certain characters incorrectly. Therefore, we opted to pretrain the model on larger databases to attain a reasonable accuracy. The model was first pretrained on two large synthetic databases, SynthText [30] (6975K) and MJSynth [31] (7224K) for 8 epochs. Then, the model is warm-started using the weights from the pretraining on the synthetic databases, and further pretrained for 120K iterations on a mixed database of large-scale real text databases OpenVINO [32] (1912K) and TextOCR [33] (710K), the previously mentioned generated synthetic database adhering to Indian license plate format (111K), another synthetic database of Indian license plate format found on Kaggle [34] (18K), and the presented database of 2237 real license plate images adhering to Indian license plate format. The vertical image concatenation data augmentation strategy was only employed on the non-license plate databases. These pretrained weights were finally finetuned for 20K iterations, exclusively on the 2237 real license plate images and a Kaggle database [20] (449 images that were labelled by us), and validated on 239 Indian license plate images from the presented database. The training was executed on an NVIDIA GeForce RTX3080 using the adam optimizer, and the One Cycle Learning Rate policy as described in [29], with a maximum learning rate of $1 \times 10^{-3}$ with a batch size of 92. The last 25% of the training steps were utilised for training the model using Stochastic Weight Averaging [35], with a learning rate of $1 \times 10^{-4}$, to improve generalisation.

TABLE III
RESULTS OF PARSEQ 224x224 WITH PATCH SIZE OF 16x16 USING
DIFFERENT DATABASE AND AUGMENTATION STRATEGIES

| | | |
|---|---|---|
| MJSynth | ./ | ./ |
| SynthText | ./ | ./ |
| OpenVINO | ./ | ./ |
| TextOCR | ./ | ./ |
| Synthetic Indian License Plates | x | ./ |
| Real Indian License Plates | ./ | ./ |
| Kaggle Indian License Plates Dataset | ./ | ./ |
| Vertically Concatenate Augmentation | x | ./ |
| Validation accuracy % | 73.13 | 95.82 |
| Validation NED | 95.70 | 99.52 |

The accuracy metric is defined as the number of correct recognitions for the entire license plate text label. NED is the Normalised Edit Distance metric for the model's predictions.

## IV. CONCLUSION

A novel database on Indian commercial vehicles' license plates was prepared and properly annotated. State-of-the-art models for real-time object detection (YOLOv7) and scene text recognition (PARSeq) were trained using this database. A license plate detection F1-score of 0.95 and a mAP0:0.5 of 0.987 was achieved, including several occluded plates, and an F1-score of 0.99 when tested only on fully visible and readable plates. Several pretraining strategies and their efficacy for this database were evaluated. Other databases and augmentation strategies were incorporated to get a more robust model with a higher validation recognition accuracy of 95.82%.

## ACKNOWLEDGMENT

The authors would like to acknowledge support by Mr. Vijay Movalia, Imagic Solutions, Ahmedabad based weighbridge service provider, for providing dataset images, and also acknowledge Mr. Dhruv R. Kabariya, Mr. Jitesh Parmar, Mr. Deep Patel, Mr. Abhi D. Patel, Ms. Kairavi R. Shah, Ms. Kavya R. Patel for assisting with data annotations and labelling.